\documentclass{article}

\usepackage{arxiv}

\usepackage[utf8]{inputenc} 
\usepackage[T1]{fontenc}    
\usepackage{hyperref}       
\usepackage{url}            
\usepackage{booktabs}       
\usepackage{amsfonts}       
\usepackage{nicefrac}       
\usepackage{microtype}      
\usepackage{lipsum}		
\usepackage{graphicx}
\usepackage[round]{natbib}
\usepackage{doi}

\usepackage{gensymb}

\usepackage{array,multirow,makecell}
\usepackage{fancybox}
\usepackage{colortbl}
\usepackage{float}

\usepackage{tikz}
\usepackage{multicol}
\usepackage{multirow}
\usepackage{booktabs}

\usepackage{algpseudocode}
\usepackage[ruled,vlined,norelsize]{algorithm2e-ali}
\usepackage{algorithmicx}

\usepackage{balance}

\usepackage{amsmath} 
\usepackage{amssymb}  

\usepackage{subfig}

\usepackage[percent]{overpic}
\usepackage{xcolor}
\usepackage{cases}

\renewcommand{\intextsep}{0pt} 

\allowdisplaybreaks 

\def\aujour{\number\day \space \ifcase\month\or
janvier\or f�vrier\or mars\or avril\or mai\or
juin\or juillet\or ao�t\or septembre\or octobre\or
novembre\or d�cembre\fi \space \number\year}

\def\cH{{\cal H}}

\def\cL{{\cal L}}

\newtheorem{remark}{Remark}

\def\C{{\setbox0=\hbox{$\displaystyle{\rm C}$}
        \hbox{\hbox to0pt{\kern 0.4\wd0\vrule height 0.95\ht0\hss}\box0}}}
\def\Q{{\setbox0=\hbox{$\displaystyle{\rm Q}$}%
    \hbox{\raise 0.2\ht0\hbox to0pt{\kern 0.4\wd0\vrule height
    0.85\ht0\hss}\box0}}} 

\def\cH2{{\cal H}_2} 

\def\cL2{\mathop{\mathcal L}_{2}} 

\def\cRH2{\mathop{\cal R \cal H}_2} 
\def\cRL2{\mathop{\cal R \cal L}_{2}} 

\makeatletter
\DeclareRobustCommand\sfrac[1]{\@ifnextchar/{\@sfrac{#1}}
                                            {\@sfrac{#1}/}}
\def\@sfrac#1/#2{\leavevmode\kern.1em\raise.5ex
         \hbox{$\m@th\fontsize\sf@size\z@
                           \selectfont#1$}\kern-.1em
         /\kern-.15em\lower.25ex
          \hbox{$\m@th\fontsize\sf@size\z@
                            \selectfont#2$}}
\makeatother

\usepackage{epstopdf}
\usepackage{rotating}
\usepackage{units}
\usepackage{siunitx}
\usepackage{float}

\title{Spatial Distribution Patterns of Clownfish in Recirculating Aquaculture Systems}

\author{ 
\href{https://orcid.org/0000-0002-3297-7778}{\includegraphics[scale=0.06]{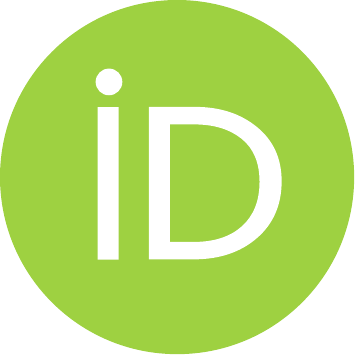}\hspace{1mm}F.~Aljehani$^1$},
\href{https://orcid.org/0000-0002-2576-5515}{\includegraphics[scale=0.06]{orcid.pdf}\hspace{1mm}I.~N'Doye$^1$}, 
\href{https://orcid.org/0000-0003-1063-8115}{\includegraphics[scale=0.06]{orcid.pdf}\hspace{1mm}M. S.~Justo$^2$}, 
\href{0000-0002-4273-2635 }{\includegraphics[scale=0.06]{orcid.pdf}\hspace{1mm}J.-E.~Majoris$^2$}, 
\href{https://orcid.org/0000-0003-2463-2742}{\includegraphics[scale=0.06]{orcid.pdf}\hspace{1mm}M.-L. Berumen$^2$}, 
\href{https://orcid.org/0000-0001-5944-0121}{\includegraphics[scale=0.06]{orcid.pdf}\hspace{1mm}T.-M.~Laleg-Kirati$^1$}
\thanks{This work has been supported by the King Abdullah University of Science and Technology (KAUST), Baseline Research Fund (BAS/1/1627-01-01) to Taous Meriem Laleg, Baseline Research Fund (BAS/1/1010-01-01) to Michael L. Berumen, and Baseline Research fund KAUST-AI Initiative Fund.} \\
$^1$Computer, Electrical and Mathematical Sciences and Engineering Division (CEMSE)\\
King Abdullah University of Science and Technology (KAUST)\\
Thuwal 23955-6900, Saudi Arabia \\
$^2$Red Sea Research Center, Biological and Environmental Science and Engineering Division\\ King Abdullah University of Science and Technology (KAUST)\\ Thuwal 23955-6900, Saudi Arabia.\\
	\texttt{\small fahad.aljehani@kaust.edu.sa; ibrahima.ndoye@kaust.edu.sa};\\\texttt{ \small john.majoris@kaust.edu.sa; micaelasofia.dossantosjusto@kaust.edu.sa;} \\
	\texttt{ \small michael.berumen@kaust.edu.sa; taousmeriem.laleg@kaust.edu.sa} \\
}

\renewcommand{\shorttitle}{\textit{arXiv} Template}

\hypersetup{
pdftitle={A template for the arxiv style},
pdfsubject={q-bio.NC, q-bio.QM},
pdfauthor={David S.~Hippocampus, Elias D.~Striatum},
pdfkeywords={First keyword, Second keyword, More},
}

\begin{document}
\maketitle

\begin{abstract}
Successful aquaculture systems can reduce the pressure and help secure the most diverse and productive Red Sea coral reef ecosystem to maintain a healthy and functional ecosystem within a sustainable blue economy. Interestingly, recirculating aquaculture systems are currently emerging in fish farm production practices. On the other hand, monitoring and detecting fish behaviors provide essential information on fish welfare and contribute to an intelligent production in global aquaculture. This work proposes an efficient approach to analyze the spatial distribution status and motion patterns of juvenile clownfish \textit{(Amphiprion bicinctus)} maintained in aquaria at three stocking densities (1, 5, and 10 individuals/aquarium). The estimated displacement is crucial in assessing the dispersion and velocity to express the clownfish's spatial distribution and movement behavior in a recirculating aquaculture system. Indeed, we aim to compute the velocity, magnitude, and turning angle using an optical flow method to assist aquaculturists in efficiently monitoring and identifying fish behavior. We test the system design on a database containing two days of video streams of juvenile clownfish maintained in aquaria. The proposed displacement estimation reveals good performance in measuring clownfish's motion and dispersion characteristics leading to assessing the potential signs of stress behaviors. We demonstrate the effectiveness of the proposed technique for quantifying variation in clownfish activity levels between recordings taken in the morning and afternoon at different stocking densities. It provides practical baseline support for online predicting and monitoring feeding behavior in ornamental fish aquaculture. As a result, we note from the spatial distribution patterns that the individual clownfish expresses a higher stress level and prefer staying at a particular location than the group dispersion and motion responses. Hence, the stocking density has a significant effect on fish behavior.
\end{abstract}

\keywords{Clownfish \and recirculating aquaculture systems (RAS) \and Optical flow method \and Displacement estimation \and Motion and dispersion behaviors \and Individual and group behaviors.}

\section{Introduction}
\textcolor{black}{The worldwide trade in aquatic ornamental species is growing rapidly over the past decade exceeding billions dollars annually \cite{Larkin2001,Wabnitz2003}}. Whereas, advanced technological systems have made aquariums much easier to monitor ornamental fish, resulted in increasing global demand \cite{SALES2003533,Chan2000}. 
Recently, model-based and model-free control strategies have been developed to track the fish growth trajectory efficiently and have shown great potential for shaping the life history of fishes and achieving better control of the output of an aquaculture system \cite{CNMBL:21,CNMBL:22}. With the development of image processing and machine vision methods in the aquaculture ornamental based industry, attention has been paid to fish health and welfare. Detecting and identifying abnormal fish behaviors are crucial to monitoring fish welfare and improving intelligent production levels in aquaculture ornamental based industry~\cite{Ash:07}. 

Changes in behavior can provide essential information regarding the health status of fish. Environmental changes that affect fish physiology and welfare are often reflected in the behavior of cultured fish \cite{LFGRC:11,KVFM:12,KPB:19,MKSLSBH:20}. Additionally, the growth performance and long-term feeding behavior are affected by the stress induced by repeated factors, and extended periods of stress can affect fish hormone levels, health, and welfare \cite{AEHB:12,SaC:16,YWAW:21}. Hence, predicting and monitoring stress behavior would be a key feature in enhancing fish welfare and production. Farmers must optimize the aquaculture system's practices, protocols, and management to guarantee optimal feeding behavior, fish growth, and monitoring throughout the grow-out cycle from stocking through harvesting \cite{Seg:16}. Thus, there is a pressing need to develop new aquaculture techniques that use recent advances in camera technology to predict and monitor spatial patterns and optimize feeding and water quality control efficiently \cite{NGPKB:00}. This objective is achieved by extracting relevant fish behavior features from video images and optimizing factors that strongly influence fish growth, such as the feeding rate, temperature, and dissolved oxygen. 

Live fish recognition is one of the most important elements in camera-based fisheries survey systems \cite{LRSXZ:03,LSSXZ:04,SGDCFN:10, YBF:14,YLH:18}. Object recognition has been well studied in various contexts in image processing and computer vision communities. However, there are fundamental challenges in identifying live fish in aquaculture systems, such as poor control over illumination, low image quality caused by fast attenuation of light in the water, and ubiquitous organic debris \cite{ YLH:18}. On the other hand, there is high uncertainty in many of the data for freely-swimming fish while capturing images due to low image quality, non-lateral fish views, or curved body shapes \cite{ YLH:18}. Fish share a strong visual correlation among species, even without uncertainty. Therefore, successfully extracting spatial distribution features is key to improving fish recognition performance and decision-making in the ornamental fish aquaculture environment.

Camera-based solutions are increasingly used for manual monitoring of fish feeding or behavior or over water for general surveillance, thanks to their ability to record accurate and continuous data \cite{JSRB:18,AKVAGroup:18,MaS:03}. The quantification of fish welfare-related metrics, such as detecting splashes'  \cite{JRBSS:16} and gill damage \cite{JSRB:18}, have recently been investigated using cameras. Video and image processing algorithms have been used for the extraction of features \cite{SSSMS:19}. Despite the noticeable progress for video/image-based fish detection and characterization, this topic is still under investigation \cite{SSSMS:19}. Additionally, there is a lack of quantitatively modeling and describing the fish movement patterns and behavioral responses.
 
Thanks to the recent advance of computer vision techniques which provide efficiently, and automatic tools for recording and analyzing specific fish behaviors, recent works have been addressed on behavioral analysis of individual fish \cite{XLCM:06,BJZG:19,ZBZZLLSY:18,GLDG:06,PPLGK:12}. However, it is worth mentioning that the fish group behaviors indicate more information about their activity level than the individual behaviors \cite{CWBSS:11,WFF:19,KEYGY:19}. Thus, it is relevant to investigate the fish group behavior in ornamental fish aquaculture.

Optical flow and kinetic energy models are typical methods used to evaluate moving objects' displacement, velocity, magnitude, and turning angle metrics. Optical flow is a powerful tool for predicting species' spatial distribution and identifying their environmental drivers' relative importance across landscapes. It detects the variations of the pixels through the image frames that extract the direction vector and velocity of objects and then calculates the motion that helps identify behaviors \cite{YWAW:21,ABF:05}. The optical flow method has been used in aquaculture to detect the particular behaviors of fish schools. Recently, spatial distribution images and optical flow methods using a convolutional neural network model to identify fish behaviors have been proposed in \cite{HZLZX:20}. \textcolor{black}{Besides, kinetic energy models can extract essential motion features to depict the motion status to detect abnormal activities in crowded (see, \cite{Zhong2007,Cao2009}). For instance, an improved version of kinetic energy models using statistical techniques combined with the optical flow method has been proposed to recognize the scattering and gathering behaviors in \cite{ZGSLLSYZ:16}}. Salmon feeding behavior has been identified through a spatio-temporal recurrent network \textcolor{black}{with significant accuracy} in \cite{MAM:19}. Displacement and velocity measurements are usually used to quantify and analyze fish feeding behavior and the movement intensity of fish populations (see, for instance, \cite{DRO:09,LLFLLL:14,ZGSLLSYZ:16,SEFPC:16}). The key analysis is based on the changes in the intensity of consecutive frames' differences induced by the fish movement patterns. 

The proposed spatial distribution algorithm based on optical flow estimation will help improve the feeding and environmental factors feedback control policies based on the finite set of fish behavior situations. For instance, the feeding, dissolved oxygen, and temperature controllers may often only need to identify the current fish behaviors to apply an action. Further, the proposed conventional optical flow estimation algorithm can relate the closed-loop behavior property with a lower computational cost than other techniques like the neural network's algorithms, which can be computationally expensive. To the best of our knowledge, no low complexity, reliable, and low-cost estimation algorithms in the literature resulting in the spatial distribution and movement behavior patterns for clownfish \textit{(Amphiprion bicinctus)} species in a recirculating aquaculture system lead to assessing the stress behaviors at different stocking densities in real-time.

In this work, we propose an estimation method of the fish displacement based on the optical flow to compute the velocity, magnitude, and turning angle for identifying the spatial distribution and motion behavior of juvenile Red Sea clownfish \textit{(Amphiprion bicinctus)} in ornamental fish aquaculture. Clownfish are commonly cultured for distribution in the marine aquarium trade, and provide an exciting model for studying the ecology and evolution of coral reef fishes because of their widespread distribution, diversity, and symbiosis with anemones \cite{BMTB:21}. Previous studies investigated on the phylogenetic relationship between clownfish and anemones to understand the evolutionary processes, the rank changes and growth rate of social hierarchies \cite{TITUS2019106526,Fitzgerald2022,BMTB:21}. Successful ornamental fish aquaculture will reduce pressure on the most productive and diverse Red Sea coral reefs to maintain a healthy and functional ecosystem. Hence, clownfish species serve as an appropriate model species for assessing the utility of optical flow for quantifying fish movement behaviors under aquaculture conditions. \textcolor{black}{We conducted two tests on the individual} and group behaviors versus morning and afternoon behaviors on a dataset containing two days of video streams of clownfish species in the Coastal and Marine Resources Lab (CMR) at King Abdullah University of Science and Technology (KAUST). Furthermore, the proposed displacement estimation performs well in identifying clownfish's motion and dispersion characteristics. \textcolor{black}{To this end, the methods and algorithms developed in this paper for clownfish species serve as a technology test in a non-replicated pilot study of clownfish behavior. Further, these test technology-oriented tools help provide practical baseline support for online predicting and monitoring particular behaviors in ornamental fish aquaculture.}


\section{Materials}\label{Materials-methods}

\subsection{Experimental system}\label{lab_experimental}
\begin{figure}[!ht]
\centering
      \begin{overpic}[scale=0.6]{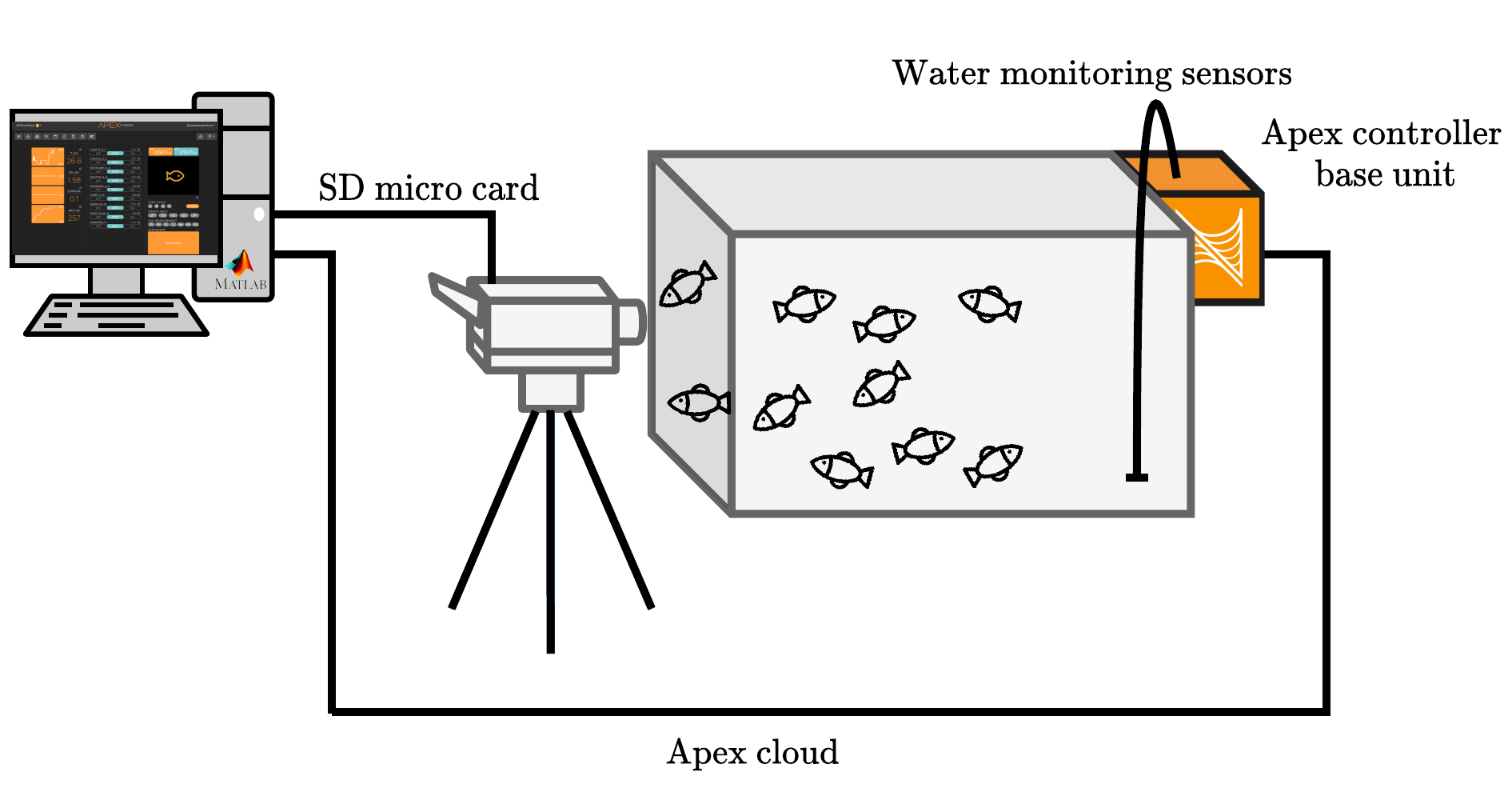}
      \put(50,39){\scriptsize Tank}
    \put(23,26){\scriptsize Camera}
    \put(76,22){\scriptsize Oxygen supply}
      \end{overpic}\vspace{-0.1cm}
      \caption{Layout of the experimental setup system}\label{Fig-1} 
 \end{figure}

A total of 16 aquacultured, juvenile clownfish \textit{(Amphiprion bicinctus)} ($0.3775\pm 0.12$ g) were maintained in three $80$ L tanks (dimension $56\times38\times45$ cm (L$\times$W$\times$H)) in the Coastal and Marine Resources Lab (CMR) at King Abdullah University of Science and Technology (KAUST), and provided with a consistent light/dark photoperiod ($12\mbox{h}$/$12\mbox{h}$). Each tank was stocked with one, five, or ten individuals to observe differences in the movement behavior of fish stocked at different densities. The tanks were equipped with an aquarium heater, air stone, and Apex Controller Base Unit that monitored and regulated the pH and temperature. A camera (Wyze Cam v3) was attached to one side of each tank, as illustrated in Fig.~\ref{Fig-1}. The cameras continuously recorded video, and the videos were saved using Micro-SD cards embedded in the cameras.  
 
During the experiment, the fish were fed twice a day ($8:00-9:00$\mbox{AM} and $15:00-16:00$\mbox{PM}). The temperatures, pH levels, and oxygen levels deviate within a certain tolerance level in the case of individual and group analysis. The maintained conditions are described as follows
\begin{itemize}
    \item For individual: Temperature at $26.07 \pm 0.19$ C$^{\circ}$, pH level at $8.04 \pm 0.04$, and oxygen levels (\%) $95.73 \pm 0.41$
    \item For group: Temperature at $26.69 \pm 0.10$ C$^{\circ}$, pH levels at $8.06 \pm 0.01$, and oxygen level (\%) $97.96 \pm 0.36$.\\
\end{itemize}
\subsection{Data preparation}
\begin{figure}[!t]
\centering
      \begin{overpic}[scale=0.5]{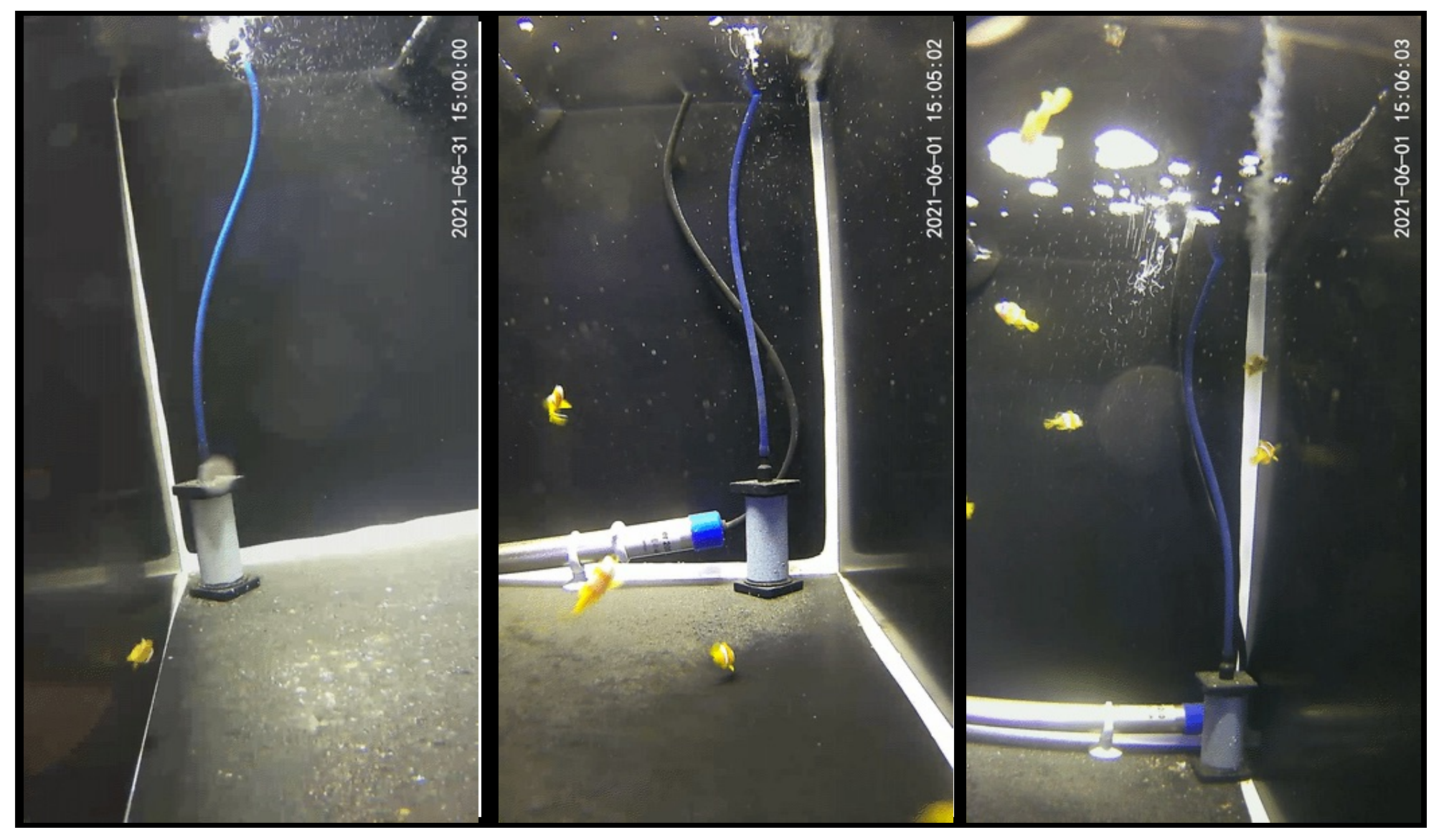}
\put(15,-3){\scriptsize\textbf{(a)}}
    \put(48,-3){\scriptsize\textbf{(b)}}
    \put(80,-3){\scriptsize \textbf{(c)}}
      \end{overpic}\vspace{0.3cm}
      \caption{KAUST Coastal and Marine Resources Lab experimental setups:  (a) one clownfish, (b) five clownfish, and (c) ten clownfish.}\label{Fig-2} 
 \end{figure}

The data collection has been performed in the light cycle for the three tanks, as illustrated in Fig.~\ref{Fig-2}. The location of the three cameras is fixed during the experiments. Various trials have been conducted to spot the best area capturing most fish movements. In Fig.~\ref{Fig-2}(a), the camera is placed at the bottom-left of the tank. And the location of the cameras in Fig.~\ref{Fig-2}(b) and (c) are set at the bottom center of the tank. Besides, we transferred the saved videos in Micro-SD cards for the three tanks to a laptop (Intel Core i9, 16GB of RAM, and 512GB Solid State Drive). The software employed to process the shared videos is \textit{MATLAB\_R2019b}.   

 \section{Methods}

 \subsection{Image processing approach}
\begin{figure}[!t]
\centering
      \begin{overpic}[scale=0.4]{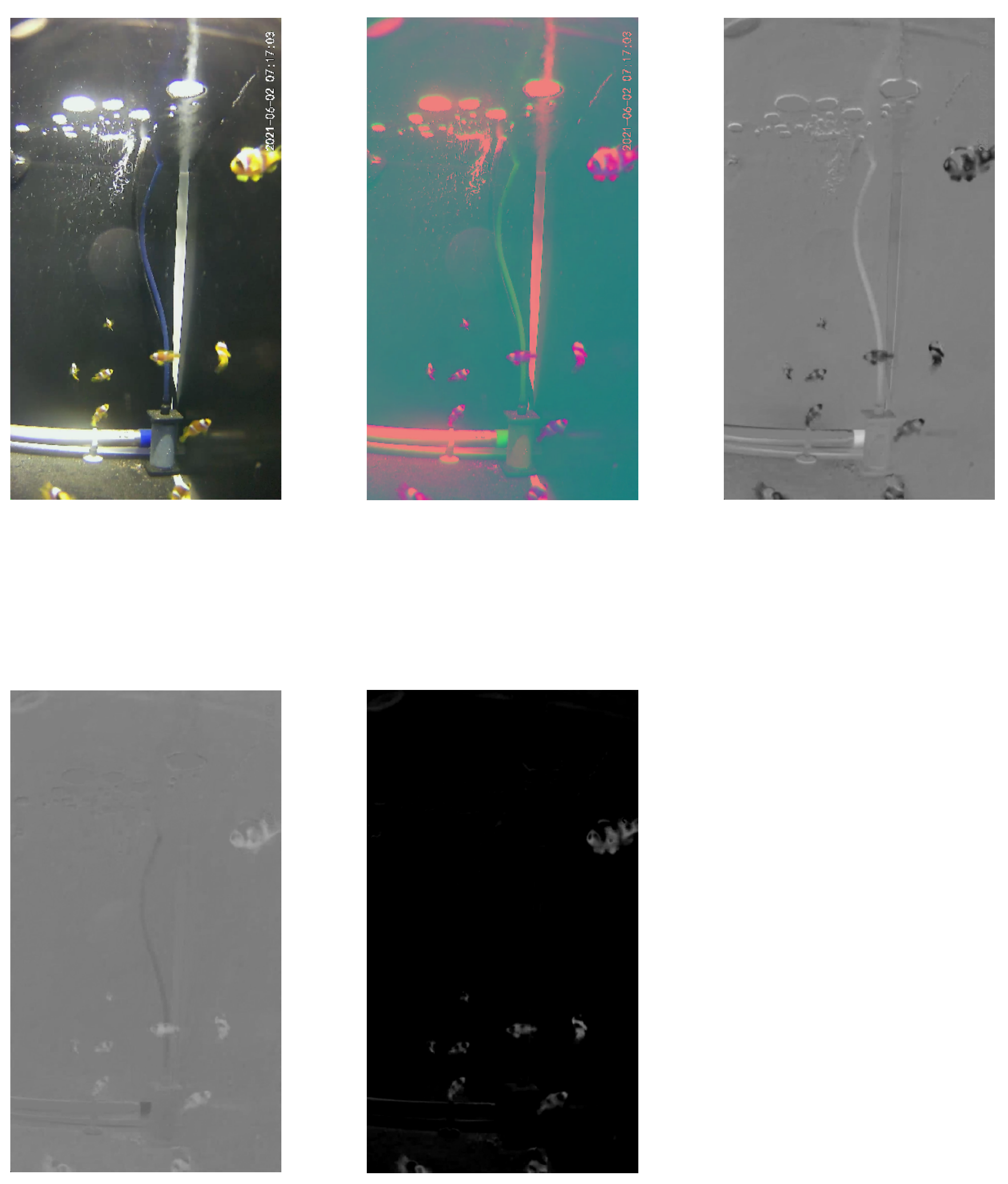}
        \put(1,100){\scriptsize\textbf{Original RGB image}}
        \put(10,53){\scriptsize\textbf{(a)}}
        \put(36,100){\scriptsize\textbf{YUV image}}
        \put(41,53){\scriptsize\textbf{(b)}}
        \put(60,100){\scriptsize\textbf{U component intensity}}
        \put(71,53){\scriptsize\textbf{(c)}}
        \put(-0.5,43){\scriptsize\textbf{V component intensity}}
        \put(10,-3){\scriptsize\textbf{(d)}}
        \put(31.5,43){\scriptsize\textbf{Difference U and V}}
        \put(41,-3){\scriptsize\textbf{(e)}}
      \end{overpic}\vspace{0.3cm}
      \caption{Color transformation from RGB to YUV components. This transformation is beneficial to remove the background and leave the clownfish.}\label{Fig-3} 
 \end{figure}
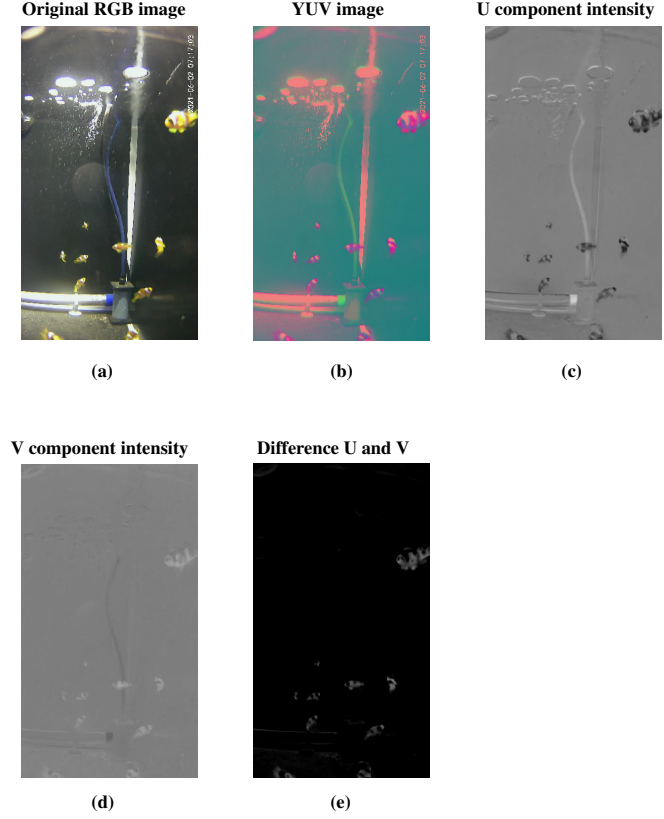

Optical flow estimates the velocity between two frames. However, the presence of oxygen bubbles caused by aquarium air stone and the light reflection  might decrease the optical flow performance. Additionally, it is challenging to differentiate between the movement of clownfish and oxygen bubbles, as shown in Fig.~\ref{Fig-2}. For this concern, we use an image processing technique to alleviate this challenge. Hence, we transfer the color intensities of all frames in RGB into YUV. Fig.~\ref{Fig-3} illustrates this transformation mapping where Fig.~\ref{Fig-3}(a) is the original image in RGB from the saved videos. Each component in RGB image is transformed to YUV as illustrated in Fig.~\ref{Fig-3}(b). Then, we extract the intensity of U and V from YUV image where Fig.~\ref{Fig-3}(c) and Fig.~\ref{Fig-3}(d) represent these extracted intensities, respectively. Finally, we significantly reduce the bubbles and light reflection by subtracting the U and V components. Fig.~\ref{Fig-3}(e) illustrates the result of this subtraction. \\

\subsection{Optical flow-based displacement estimation}\label{optical flow}
\begin{figure}[!t]
\centering
      \begin{overpic}[scale=0.58]{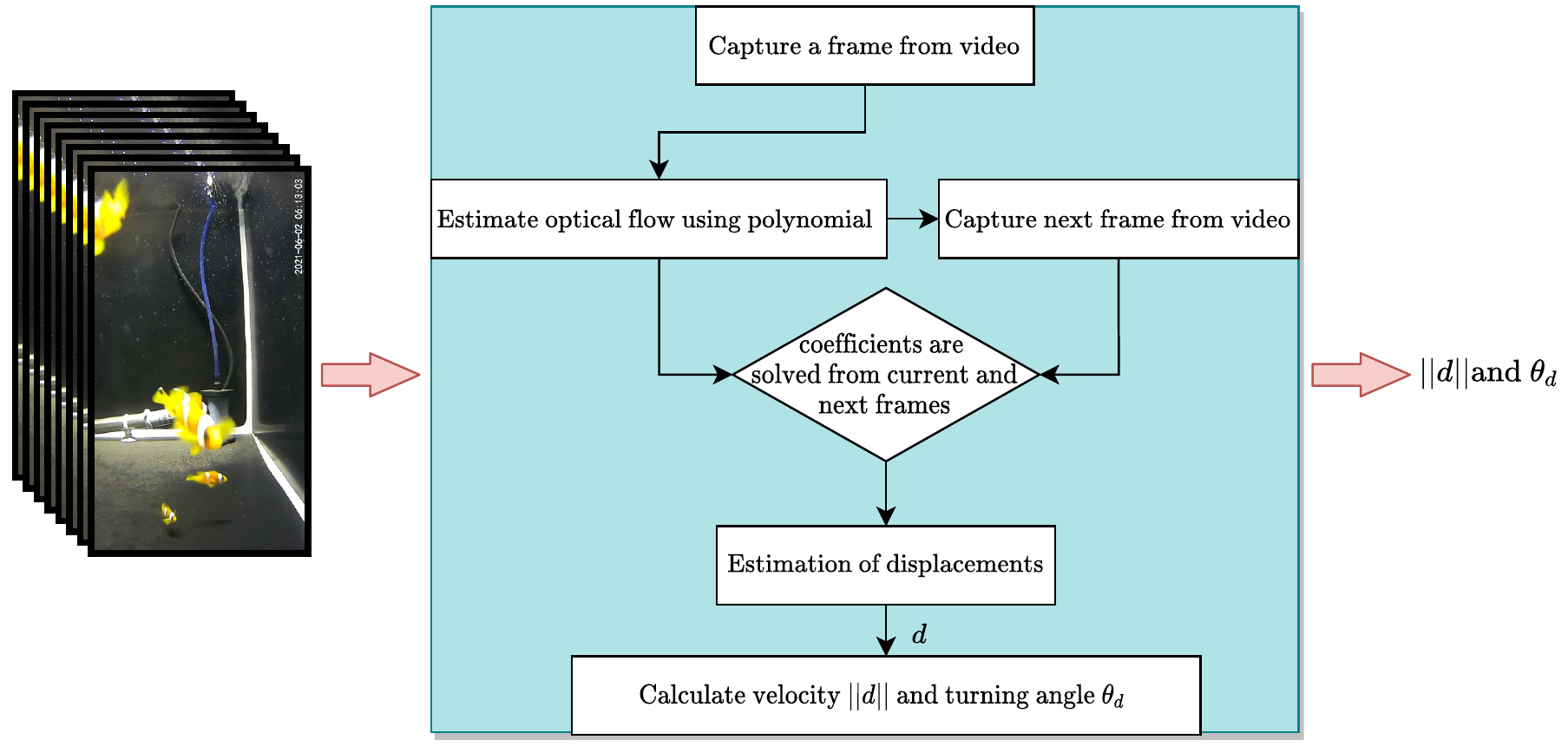}
    \end{overpic}\vspace{-0.1cm}
      \caption{Flowchart of optical flow-based displacement estimation}\label{Fig-4}  
 \end{figure}

We propose an optical flow estimation method to capture the motion of moving fish from two consecutive frames. This methodology relies on the two-frame motion estimation results \cite{Farneback}. It also assumes that the neighborhood of pixels in $(x,y)$ coordinate can be approximated by a quadratic polynomial form as follows
\begin{equation}\label{quad_image1}
    f_t(\mathcal{X}) \approx \mathcal{X}^TA_t\mathcal{X} + b_t^T\mathcal{X}+ c_t,
\end{equation}
where $\mathcal{X}$ is the coordinate $(x,y)$ of the first image at time $t$, $A_t$, $b_t$, and $c_t$ are the quadratic parameters of the first image. A new signal $f_{t+1}$ can be constructed by the motion of moving objects between two frames, $d$, as
\begin{equation}\label{quad_image2}
    f_{t+1}(\mathcal{X}) = f_t(\mathcal{X}-d),
\end{equation}
Substituting \eqref{quad_image1} into \eqref{quad_image2}, one can have the relationship between the two consecutive images as follows
 \begin{equation}\label{nextQuad}
    f_{t+1}(\mathcal{X}) = \mathcal{X}^TA_{t+1}\mathcal{X}+ b_{t+1}^T\mathcal{X}+c_{t+1}
\end{equation}
where
\begin{align*}
    A_{t+1} &= A_t\\
    b_{t+1}^T &= (b_t-2A_td)^T\\
    c_{t+1} &= d^TA_td- b_t^Td+c_t
\end{align*}
From \eqref{nextQuad}, the motion vector $d$ is  computed as follows
\begin{align}\label{d_ideal}
    d &= \frac{1}{2}A_t^{-1}(b_t-b_{t+1}).
\end{align}
Hence, the velocity and turning angle of moving fishes can be obtained as
\begin{equation}\label{equat1}
    ||d|| = \sqrt{d_x^2 + d_y^2},\qquad 
    \theta_d = \tan^{-1}{({d_y}/{d_x})},
\end{equation}
where $d_x$ and $d_y$ refer to displacement in horizontal direction $x$ and vertical direction $y$ between two frames respectively. Note that equation \eqref{d_ideal} provides the motion vector in ideal case since it assumed that $A_t=A_{t+1}$. Fig.~\ref{Fig-4} illustrates the optical flow-based displacement estimation algorithm.
 
\begin{remark}
From a practical point of view, the motion vector $d$ of  consecutive frames is given as follows 
\begin{equation}\label{motion-equa1}
    A_{\mbox{\scriptsize ave}}(\mathcal{X})d(\mathcal{X}) =  \Delta b(\mathcal{X}),
\end{equation}
with 
\begin{equation*}\label{motion-equa2}
A_{\mbox{\scriptsize ave}}(\mathcal{X}) = \frac{A_t(\mathcal{X})+A_{t+1}(\mathcal{X})}{2}, \qquad
\Delta b(\mathcal{X}) = \frac{b_t(\mathcal{X})-b_{t+1}(\mathcal{X})}{2},
\end{equation*}
where $A_{\mbox{\scriptsize ave}}(\mathcal{X})$ is the two frames average parameter.
\end{remark}

\begin{figure}[!ht]
\centering
      \begin{overpic}[scale=0.78]{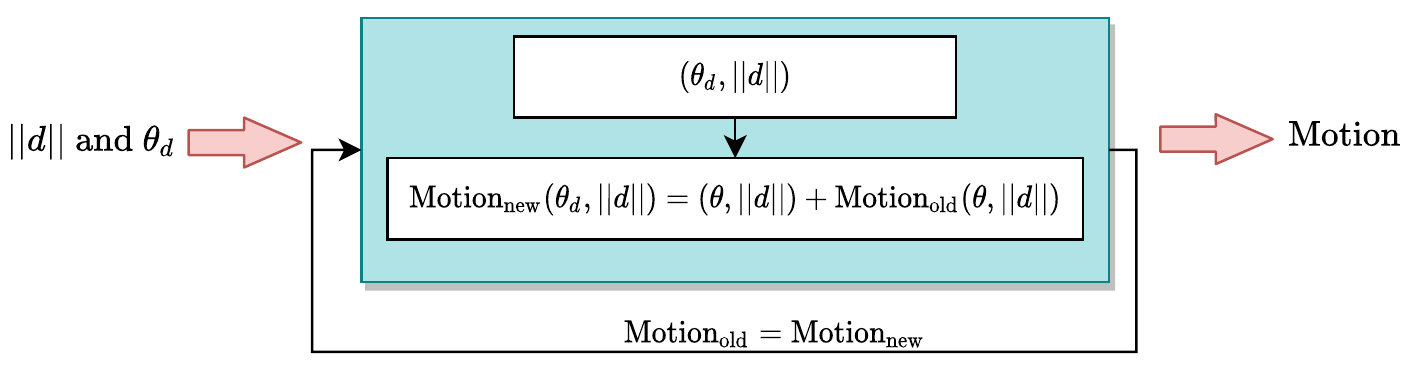}
      \end{overpic}\vspace{0.3cm}
      \caption{Motion framework}\label{Fig-6}  
 \end{figure}

The statistical analysis in this study depends mainly on dispersion status and motion characteristics to extract behavioral metrics. Using optical flow-based displacement estimation yields velocity $||d||$ and turning angle $\theta_d$. From the velocities and turning angles, statistical analysis is utilized as follows:
\begin{itemize}
\item {\bf Dispersion status steps:}
The primary goal of using the dispersion metric is to investigate the spatial behavior of the clownfish spread in the tanks. In other words, it is a visual representation that allows understanding the clownfish behavior adequately. Fig.~\ref{Fig-5} summarizes the dispersion steps where the focus is to consider the velocity level of the clownfish. The tolerance is chosen to detect movements of the clownfish.\\
 \begin{figure}[!ht]
\centering
      \begin{overpic}[scale=0.78]{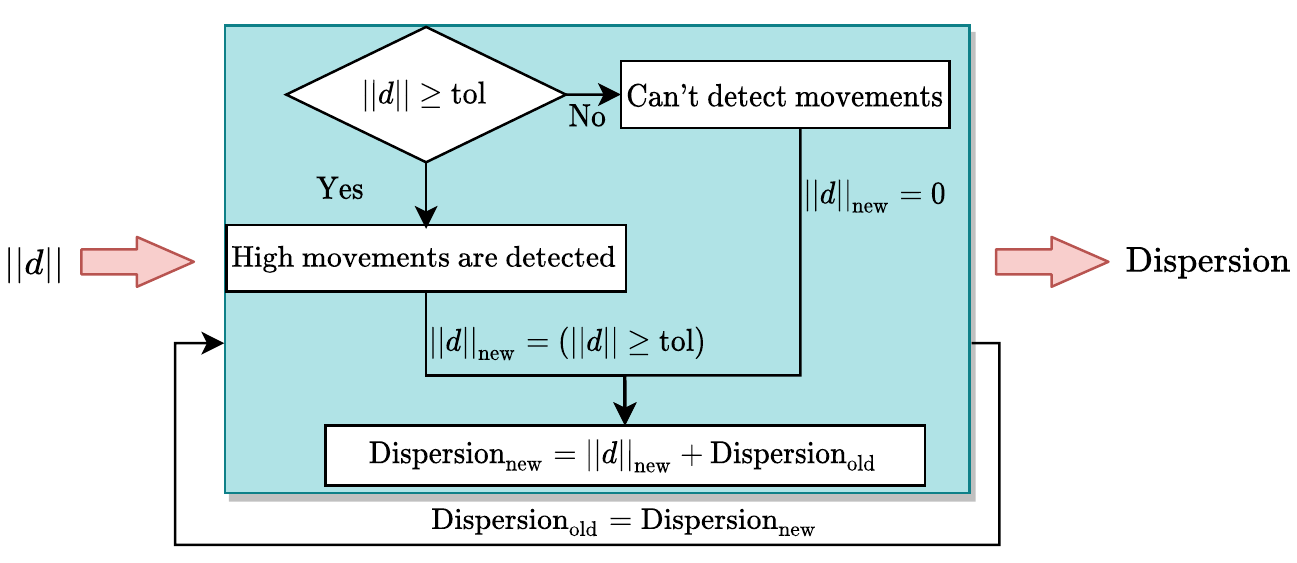}
      \end{overpic}\vspace{-0.1cm}
      \caption{Dispersion framework}\label{Fig-5}  
 \end{figure}

\item {\bf Motion characteristics steps:}
Velocity and turning angle of the moving clownfish are essential indices to quantify the stress behavior. The objective is to investigate the motion characteristic of the clownfish using their velocities and turning angles. The steps to obtain motion characteristic is illustrated in Fig.~\ref{Fig-6}.
\end{itemize}

 \begin{figure*}[!ht]
\centering
    \begin{overpic}[scale=0.40]{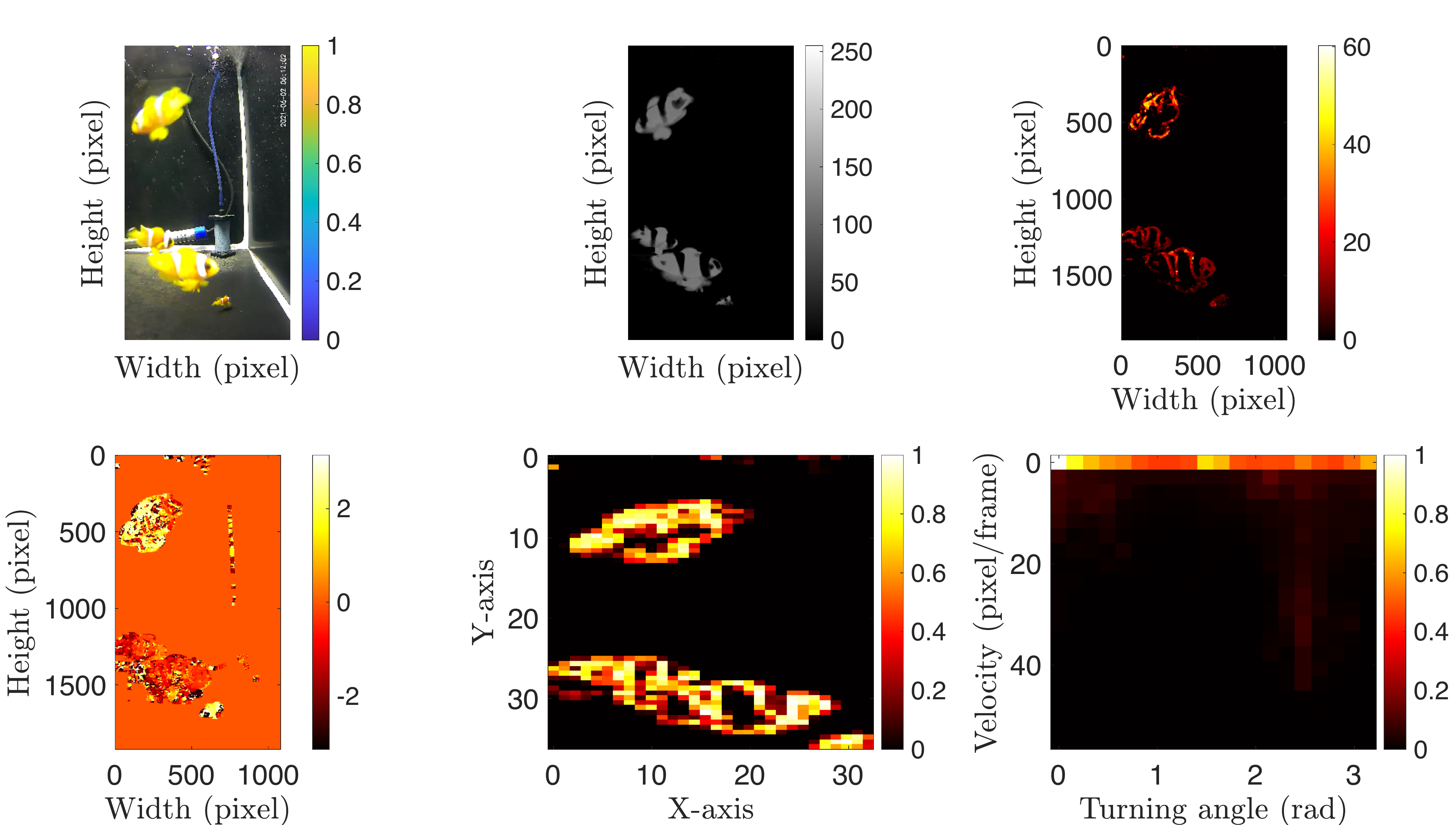}
          \put(13,55){\small\textbf{(a)}}
          \put(48 ,55){\small\textbf{(b)}}
          \put(82,55){\small\textbf{(c)}}
          \put(13,-2.3){\small\textbf{(d)}}
          \put(48,-2.2){\small\textbf{(e)}}
          \put(82,-2.2){\small \textbf{(f)}}
    \end{overpic}\vspace{0.4cm}
    \caption{The results of dispersion and motion frameworks: (a) original video, (b) the difference intensity of U and V components, (c) the magnitude of the velocity after applying the threshold, (d) the turning angle of the velocity, (e) the dispersion analysis, and (f) the motion analysis.}\label{Fig-7}
      \vspace{0.5cm}
 \end{figure*}
 
\textcolor{black}{Algorithm \ref{algo1} provides the pseudocode to implement the dispersion status and motion characteristics steps based on the optical flow method.}

\begin{algorithm-ali}
\textcolor{black}{
    \SetKwInOut{KwIn}{Input}
    \SetKwInOut{KwOut}{Output}
    \textcolor{black}{
    \KwIn{Recorded videos\\ \ \ $\mbox{Dispersion}(1)= \textbf{0}$\\ \ \ $\mbox{Motion}(1)= \textbf{0}$}
    \KwOut{Dispersion and motion histories}
    \tcc{Convert video to frames}
    Frames = VideoReader(Recorded videos) \\
    \tcc{Convert original RBG frame to YUV}
    Frames = rgb2yuv(Frames) \\
    \For{$i = 1$ \KwTo \mbox{length(Frames)}}{
    \tcc{Equations \eqref{quad_image1}-\eqref{motion-equa1}}
    $\left(||d||,\theta_d\right) = \mbox{OpticalFlow(Frames)}$ \\
        \eIf{$||d|| \geqslant tol$}{
            $||d||_{\text{new}} = ||d||$ \tcp*{motion detected}
         }{
            $||d||_{\text{new}} = 0$ \tcp*{motion not detected}
         }
         \tcc{Update dispersion status}
         $\mbox{Dispersion}(i+1) = \mbox{Dispersion}(i) + ||d||_{\text{new}} $\\
         \tcc{Update motion characteristics}
         $\mbox{Motion}(i+1) = \mbox{Motion}(i) + \left(||d||_{\text{new}},\theta_d\right)$
    }
    \KwRet{\mbox{Dispersion and motion}}
    \caption{Dispersion and motion estimation}\label{algo1}
    }
}
\end{algorithm-ali}

\section{Results and discussion}\label{section4}
This section presents the joint dispersion and motion responses for individual and group fish to quantify the clownfish activity levels through the morning and afternoon settings. Indeed, the joint dispersion status and motion characteristics allow us to extract behavioral metrics from two frames motions. Fig.~\ref{Fig-7} summarizes the overall algorithm of extracting behavioral metrics in which Fig.~\ref{Fig-7}(a) represents one frame from the recorded video. Fig.~\ref{Fig-7}(b) is the result of subtracting the U and V intensities after transferring the RGB components to YUV. The result of this subtraction is used as input to the optical flow algorithm to estimate the velocity. Fig.~\ref{Fig-7}(c) and Fig.~\ref{Fig-7}(d) are the magnitude velocity and turning angle, respectively. From the magnitude velocity and turning angle, we extract the behavioral dispersion and motion patterns. Fig.~\ref{Fig-7}(e) shows the joint dispersion status. The height and width of the original frame are projected or scaled to $36 \times 30$, where the height becomes $36$ and the weight is $30$. The black spots indicate the undetected movements or not visited places in the tank. The variations from black to white bar on the right of the figure indicate the occurring probability. Fig.~\ref{Fig-7}(f) presents the joint motion characteristics. Due to the symmetric of the turning angle in joint motion, the turning angle is plotted from $0 \text{ to } 3$ radius, instead of $-3 \text{ to } 3$ radius. \textcolor{black}{The implementation of the proposed spatial distribution algorithm based on optical flow estimation and the offline clownfish video datasets of our experiments are provided in the repository at \url{https://github.com/EMANG-KAUST/ComputerVisionFishBehavior.git}}.

 \begin{figure*}[!t]
\begin{minipage}{1\textwidth}
\centering
      \begin{overpic}[scale=0.32]{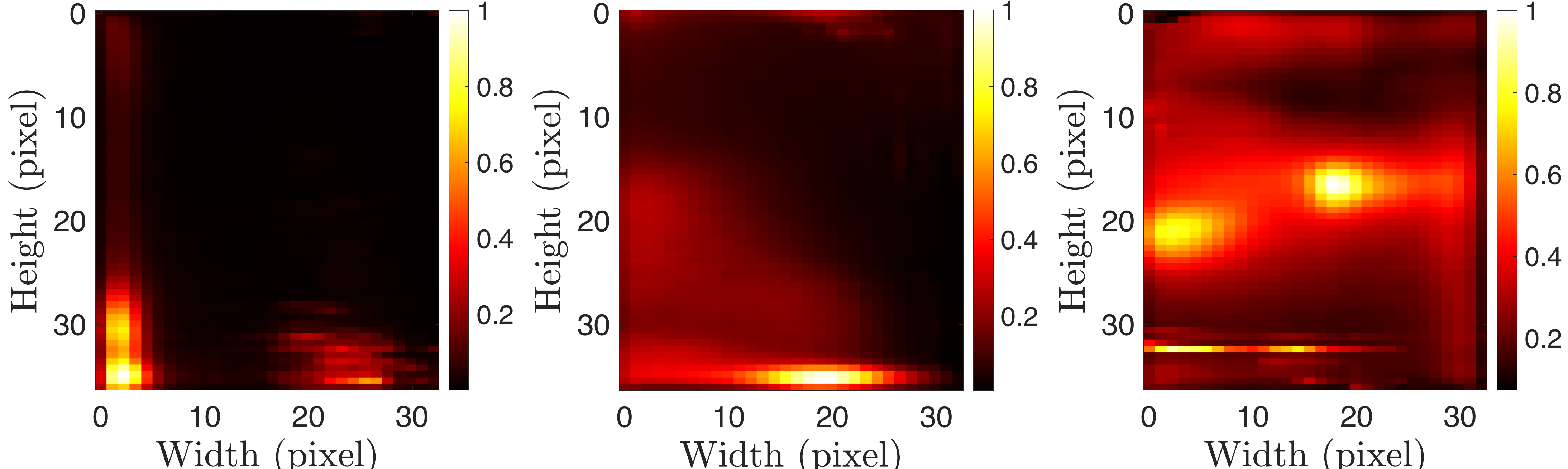}
          \put(16,-3){\scriptsize\textbf{(a)}}
          \put(49,-3){\scriptsize\textbf{(b)}}
          \put(83,-3){\scriptsize\textbf{(c)}}
          \put(12,30.5){\scriptsize\textbf{One clownfish}}
          \put(45,30.5){\scriptsize\textbf{Five clownfish}}
          \put(79,30.5){\scriptsize\textbf{Ten clownfish}}
      \end{overpic}
         \end{minipage}\vspace{0.3cm}
      \caption{Averaged dispersion on $ \text{June } 02, \ 2021$ from $7:00~\mbox{am}$ to $15:00~\mbox{pm}$. The bar next to the figures illustrates occurrence probability where the black spots present less probable visited locations and white spots present most probable visited locations.}\label{Fig-8} 
\begin{minipage}{1\textwidth}\vspace{0.3cm}
\centering
      \begin{overpic}[scale=0.32]{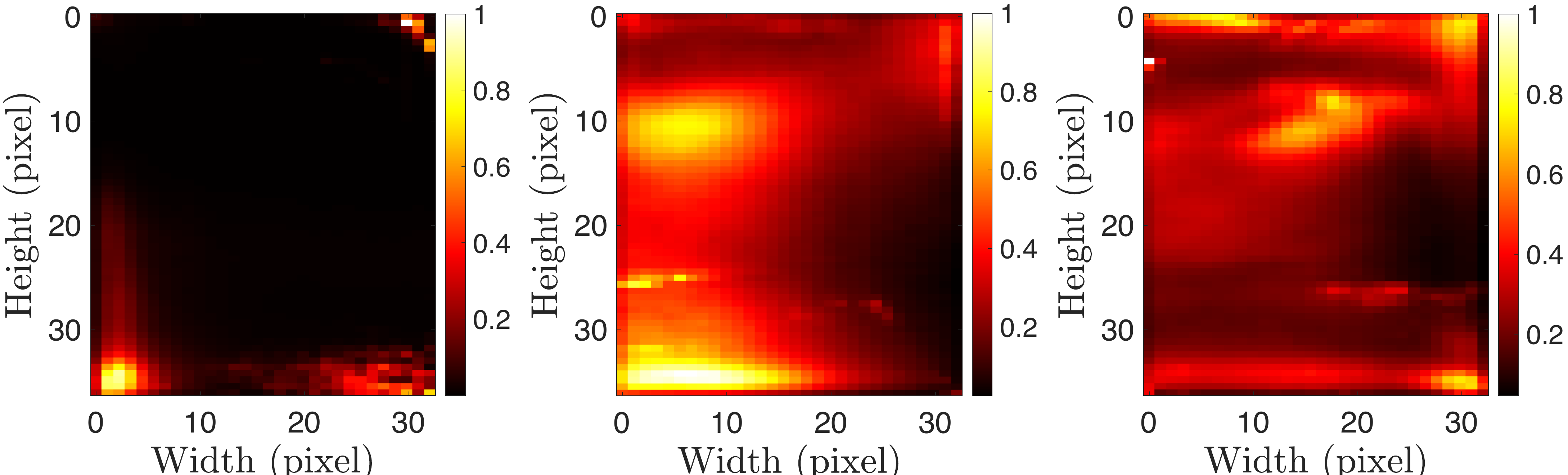}
          \put(16,-3){\scriptsize\textbf{(a)}}
          \put(49,-3){\scriptsize\textbf{(b)}}
          \put(83,-3){\scriptsize\textbf{(c)}}
          \put(12,30.5){\scriptsize\textbf{One clownfish}}
          \put(45,30.5){\scriptsize\textbf{Five clownfish}}
          \put(79,30.5){\scriptsize\textbf{Ten clownfish}}
      \end{overpic}
      \end{minipage} \vspace{0.3cm}
      \caption{Averaged dispersion on $ \text{June } 08, \ 2021$ from $7:00~\mbox{am}$ to $15:00~\mbox{pm}$.}\label{Fig-9} 
      \vspace{0.3cm}
 \end{figure*}
 
\begin{figure*}[!t]
\begin{minipage}{1\textwidth}
\centering
      \begin{overpic}[scale=0.32]{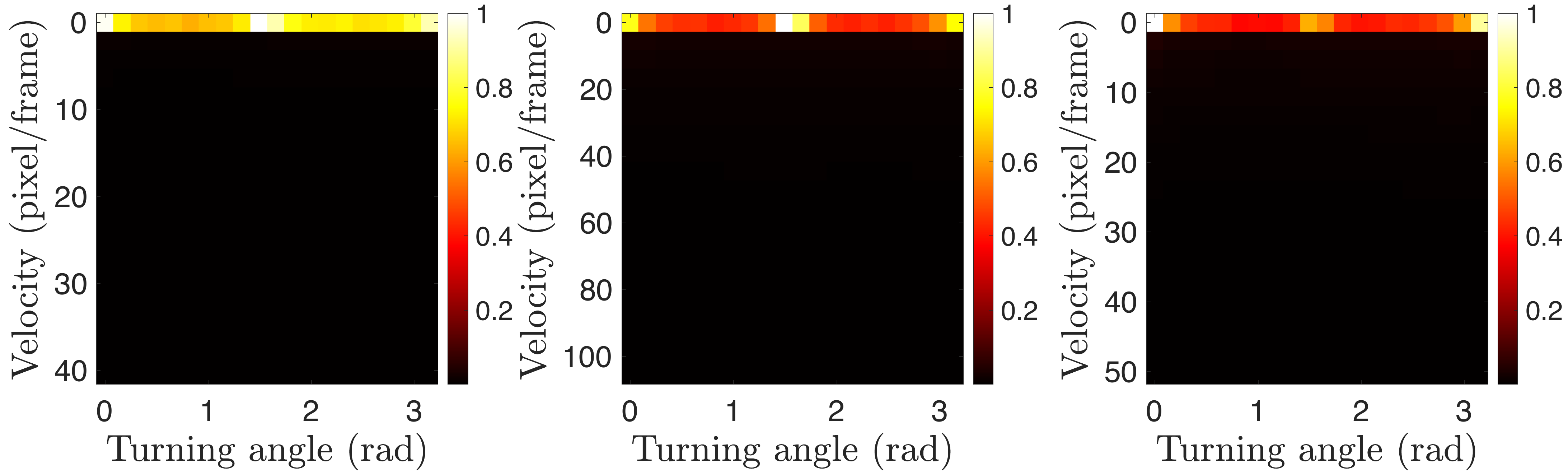}
          \put(16,-3){\scriptsize\textbf{(a)}}
          \put(49,-3){\scriptsize\textbf{(b)}}
          \put(83,-3){\scriptsize\textbf{(c)}}
          \put(12,30.5){\scriptsize\textbf{One clownfish}}
          \put(45,30.5){\scriptsize\textbf{Five clownfish}}
          \put(79,30.5){\scriptsize\textbf{Ten clownfish}}
      \end{overpic}
      \end{minipage}\vspace{0.3cm}
      \caption{Averaged motion on $ \text{June } 02, \ 2021$ from $7:00~\mbox{am}$ to $15:00~\mbox{pm}$. The figures relate the velocity of clownfish with respect to moving angle.}\label{Fig-10} 
\begin{minipage}{1\textwidth}\vspace{0.3cm}
\centering
      \begin{overpic}[scale=0.32]{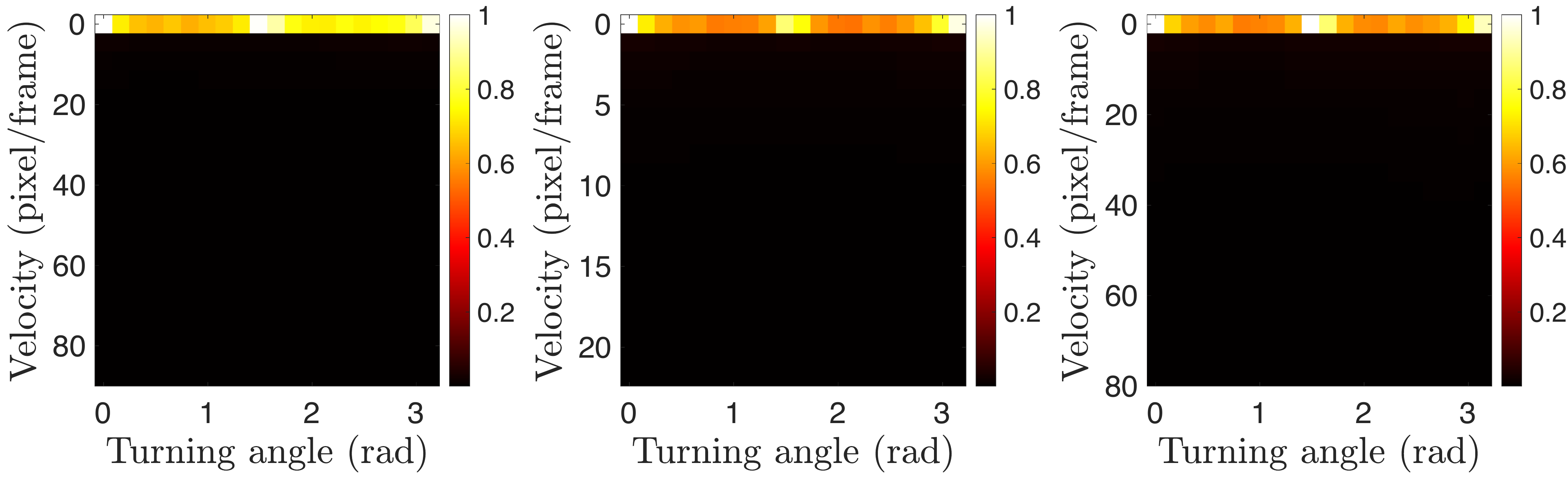}
          \put(16,-3){\scriptsize\textbf{(a)}}
          \put(49,-3){\scriptsize\textbf{(b)}}
          \put(83,-3){\scriptsize\textbf{(c)}}
          \put(12,30.5){\scriptsize\textbf{One clownfish}}
          \put(45,30.5){\scriptsize\textbf{Five clownfish}}
          \put(79,30.5){\scriptsize\textbf{Ten clownfish}}
      \end{overpic}
      \end{minipage} \vspace{0.3cm}
      \caption{Averaged motion on $ \text{June } 08, \ 2021$ from $7:00~\mbox{am}$ to $15:00~\mbox{pm}$.}\label{Fig-11} 
 \end{figure*}
 
\subsection{Individual and group analysis: dispersion and motion patterns}\label{subsection1aa}
This subsection aims to study the extracted information from individual and group clownfish using the joint dispersion status and motion characteristics. For this purpose, we perform an average test that occurs from $7\!:\!00~\mbox{am}$ to $15\!:\!00~\mbox{pm}$ to identify the joint dispersion status and motion characteristics in individual and group behaviors. Figs.~\ref{Fig-8} and \ref{Fig-9} show the joint dispersion responses within two days (on June 02, 2021, and June 08. 2021). One can observe that the number of high occurrence probabilities containing the dispersion information of the clownfish in individual behavior is less than that in group behavior. The individual clownfish stays in one preferable location and hesitates to swim all over the tank. At the same time, the group dispersion spreads almost everywhere in the tanks. Figs.~\ref{Fig-10} and \ref{Fig-11} illustrate the joint motion responses for the two days. 
\textcolor{black}{The occurrence probability of individual behavior is higher than group behavior. Indeed, this high occurrence level from the joint motion responses might be possible signs of stress in individual clownfish resulting maybe from the stocking density, environment or management factors. In addition, the individual clownfish is more sensitive to any latent signs of stress behavior and prefers staying at a particular location from the joint dispersion and motion responses. On the other hand, the group clownfish is less receptive to stress levels while swimming freely in the tanks. However, these obtained observations results related to stress behavior cannot be extrapolated to the clownfish population in general, as they do not include replicated pilots. Indeed, the underlying premise of this work is to infer a technology test in a non-replicated pilot study of clownfish behavior.}

\subsection{Joint dispersion responses: morning versus afternoon}\label{subsection1a}
\begin{figure*}[!t]
\begin{minipage}{1\textwidth}
\centering
      \begin{overpic}[scale=0.32]{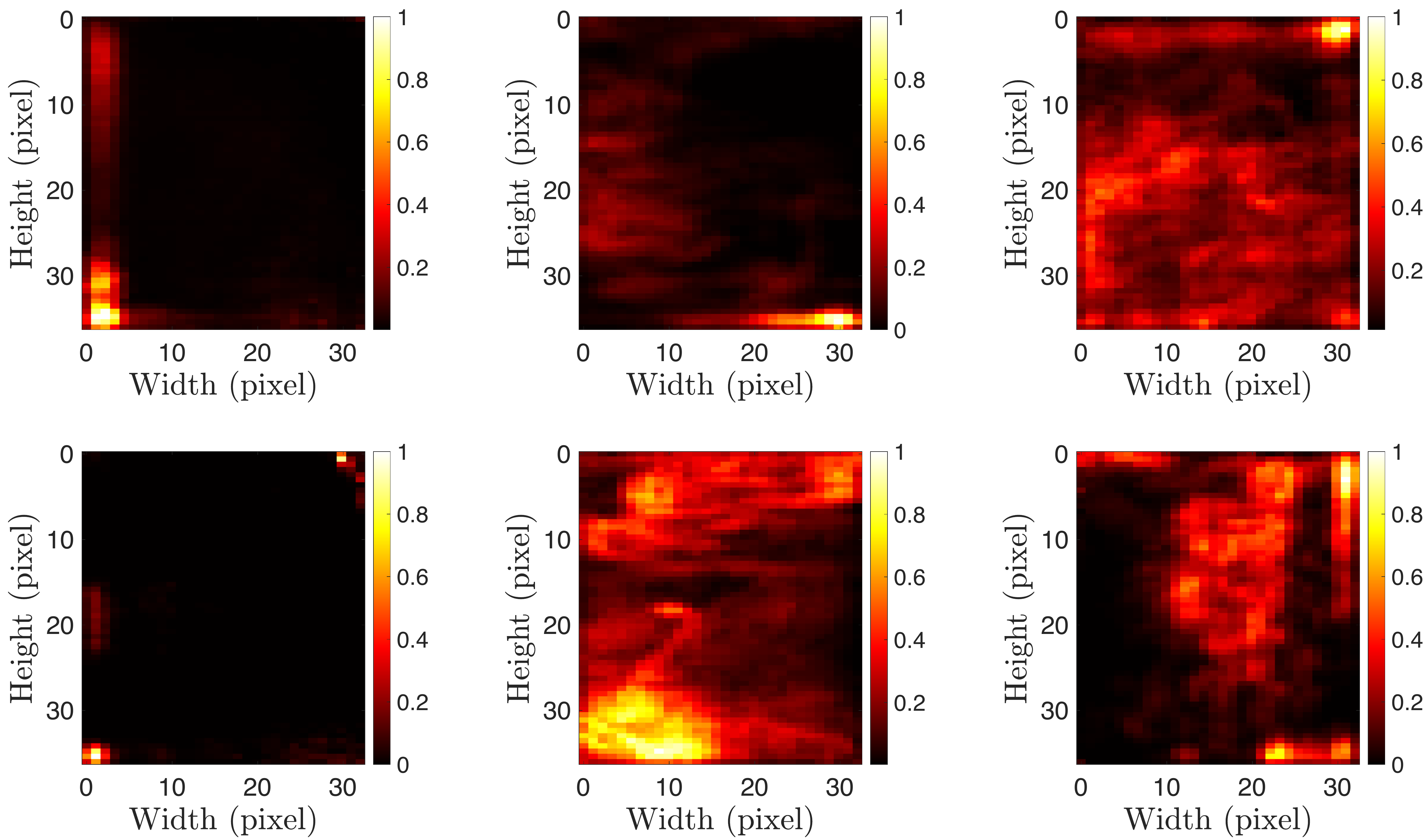}
          \put(14,28){\scriptsize\textbf{(a)}}
          \put(14,-3){\scriptsize\textbf{(d)}}
          \put(49,28){\scriptsize\textbf{(b)}}
          \put(49,-3){\scriptsize\textbf{(e)}}
          \put(83,28){\scriptsize\textbf{(c)}}
          \put(83,-3){\scriptsize\textbf{(f)}}
          \put(11,59){\scriptsize\textbf{One clownfish}}
          \put(45,59){\scriptsize\textbf{Five clownfish}}
          \put(79,59){\scriptsize\textbf{Ten clownfish}}
      \end{overpic} 
      \end{minipage}\vspace{0.3cm}
      \caption{Five minutes dispersion on $ \text{June } 02, \ 2021$ where the morning starts from $7:01~\mbox{am}$ to $7:05~\mbox{am}$ and afternoon starts from $13:01~\mbox{pm}$ to $13:05~\mbox{pm}$. Top figures illustrate morning dispersion and bottom figures present afternoon dispersion.}\label{Fig-12}       
 \begin{minipage}{1\textwidth}\vspace{0.5cm}     
      \centering
      \begin{overpic}[scale=0.34]{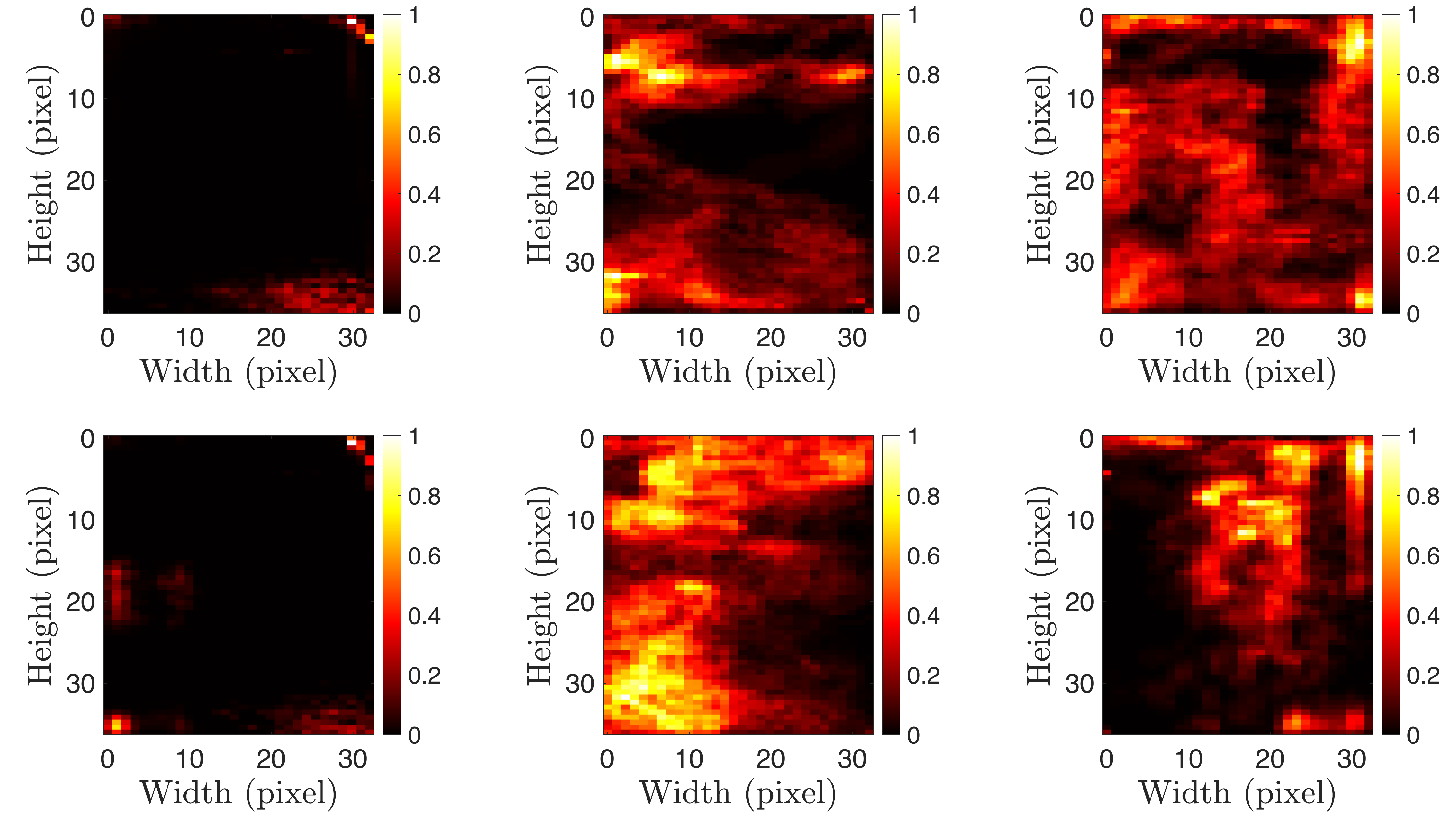}
          \put(14,28){\scriptsize\textbf{(a)}}
          \put(14,-3){\scriptsize\textbf{(d)}}
          \put(49,28){\scriptsize\textbf{(b)}}
          \put(49,-3){\scriptsize\textbf{(e)}}
          \put(83,28){\scriptsize\textbf{(c)}}
          \put(83,-3){\scriptsize\textbf{(f)}}
          \put(11,59){\scriptsize\textbf{One clownfish}}
          \put(45,59){\scriptsize\textbf{Five clownfish}}
          \put(79,59){\scriptsize\textbf{Ten clownfish}}
      \end{overpic} 
      \end{minipage} \vspace{0.3cm}
      \caption{\textcolor{black}{Five minutes dispersion on $ \text{June } 08, \ 2021$ where the morning starts from $7:01~\mbox{am}$ to $7:05~\mbox{am}$ and afternoon starts from $13:01~\mbox{pm}$ to $13:05~\mbox{pm}$. Top figures illustrate morning dispersion and bottom figures present afternoon dispersion.}}\label{Fig-13}
 \end{figure*}

We discuss the morning and afternoon cases in which the individual and group responses are evaluated using the joint dispersion status and motion characteristics metrics. 
This study consists of two days of experiments on $2^\text{nd}$  and $8^\text{th}$ June 2021. The morning case starts from $7:01~\mbox{am}$ till $7:05~\mbox{am}$ and the afternoon case begins from $13:01~\mbox{pm}$ to $13:05~\mbox{pm}$. Fig.~\ref{Fig-12} shows the joint dispersion results on $2^\text{nd}$ June within almost five minutes of recording data. Figs.~\ref{Fig-12}(a), (b), and (c) represent the morning case for one, five, and ten clownfish, respectively. On the other hand, Figure~\ref{Fig-12}(d), (e), and (f) illustrate the afternoon case for one, five, and ten clownfish. 

\textcolor{black}{We observed that the single clownfish always stayed at the bottom right of the tank during the morning and afternoon while the groups of clownfish swam everywhere in the tanks}. Additionally, the swimming path, derived from the individual's extracted joint dispersion, \textcolor{black}{was} almost predictable. Furthermore, regardless of the morning and afternoon cases, the individual \textcolor{black}{behaved} similarly, swimming at the bottom-right corner. At the same time, the swimming paths of the group \textcolor{black}{varied} in the morning and afternoon. Fig.~\ref{Fig-13} illustrates the joint dispersion results on $8^\text{th}$ June within almost five minutes of recording data. The analysis of $8^\text{th}$ June is to confirm the observations we had in $2^\text{nd}$ June. Indeed, the behaviors in the morning and afternoon of individual and group clownfish \textcolor{black}{were} the same.

\subsection{Joint motion responses: morning versus afternoon}\label{subsection2a}
\begin{figure*}[!t]
\begin{minipage}{1\textwidth}
\vspace{0.2cm}
\centering
      \begin{overpic}[scale=0.3]{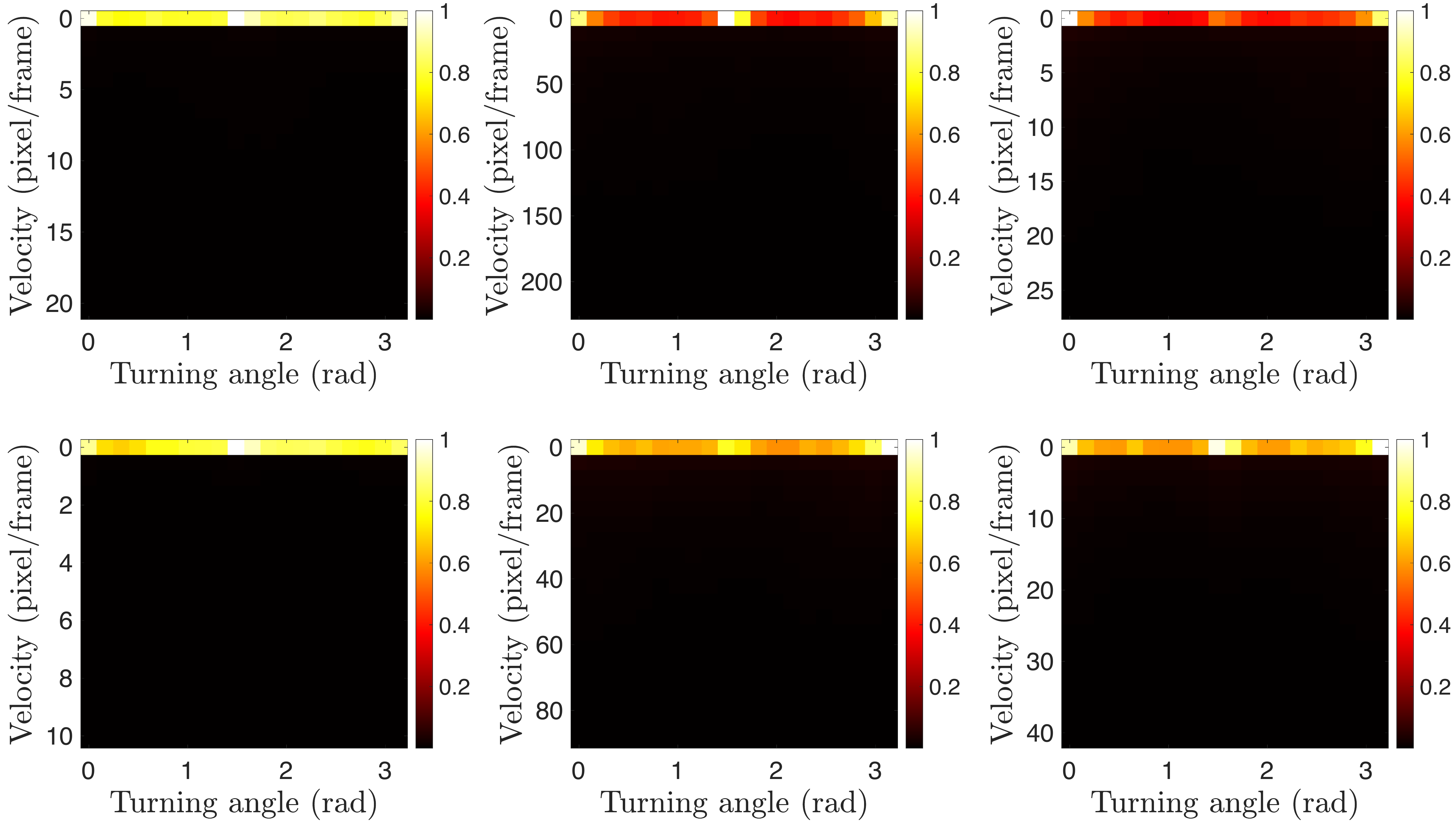}
          \put(16,27.5){\scriptsize\textbf{(a)}}
          \put(16,-2.5){\scriptsize\textbf{(d)}}
          \put(49,27.5){\scriptsize\textbf{(b)}}
          \put(49,-2.5){\scriptsize\textbf{(e)}}
          \put(83,27.5){\scriptsize\textbf{(c)}}
          \put(83,-2.5){\scriptsize\textbf{(f)}}
          \put(11,57.5){\scriptsize\textbf{One clownfish}}
          \put(45,57.5){\scriptsize\textbf{Five clownfish}}
          \put(79,57.5){\scriptsize\textbf{Ten clownfish}}
      \end{overpic} 
       \end{minipage}\vspace{0.3cm}
      \caption{Five minutes motion on $ \text{June } 02, \ 2021$ where the morning starts from $7:01~\mbox{am}$ to $7:05~\mbox{am}$ and afternoon starts from $13:01~\mbox{pm}$ to $13:05~\mbox{pm}$. Top figures illustrate morning motion and bottom figures present afternoon motion.}\label{Fig-18} 
 \begin{minipage}{1\textwidth}\vspace{1.5cm} 
\centering
      \begin{overpic}[scale=0.32]{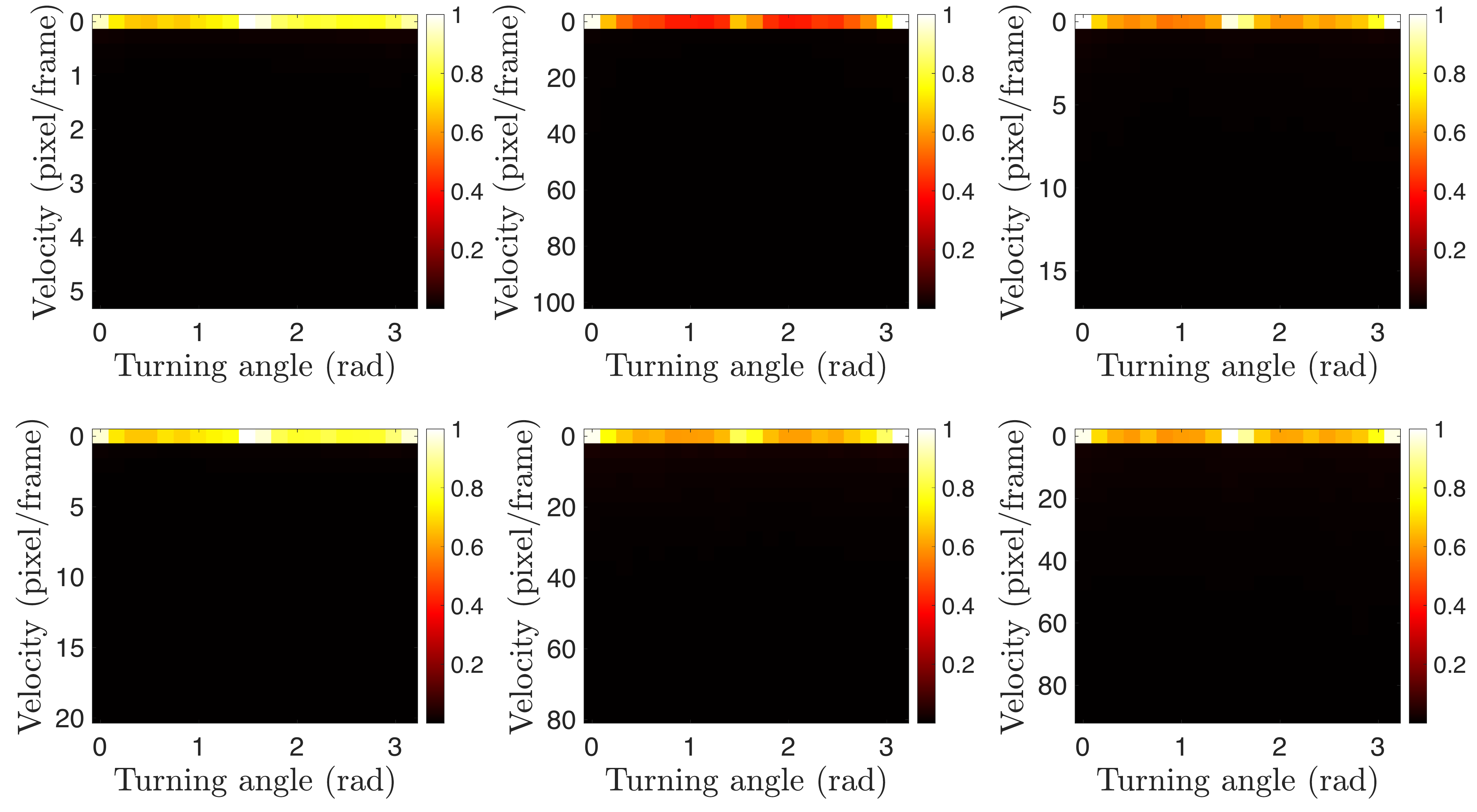}
          \put(16,27.5){\scriptsize\textbf{(a)}}
          \put(16,-2.5){\scriptsize\textbf{(d)}}
          \put(49,27.5){\scriptsize\textbf{(b)}}
          \put(49,-2.5){\scriptsize\textbf{(e)}}
          \put(83,27.5){\scriptsize\textbf{(c)}}
          \put(83,-2.5){\scriptsize\textbf{(f)}}
          \put(11,57.5){\scriptsize\textbf{One clownfish}}
          \put(45,57.5){\scriptsize\textbf{Five clownfish}}
          \put(79,57.5){\scriptsize\textbf{Ten clownfish}}
      \end{overpic} 
       \end{minipage} \vspace{0.3cm}
      \caption{\textcolor{black}{Five minutes motion on $ \text{June } 08, \ 2021$ where the morning starts from $7:01~\mbox{am}$ to $7:05~\mbox{am}$ and afternoon starts from $13:01~\mbox{pm}$ to $13:05~\mbox{pm}$. Top figures illustrate morning motion and bottom figures present afternoon motion.}}\label{Fig-19}  
 \end{figure*}

We conducted five-minute joint motion performances of the individual and group behaviors based on morning and afternoon scenarios. The motion performances are based on the velocity and turning angle factors which measure the clownfish's status. Figs.~\ref{Fig-18} and \ref{Fig-19} illustrate the five-minute motion performances that express the clownfish's spatial behavioral characteristics. In both days and scenarios, the individual occurrence velocity's probability is higher than the group behavior, which reveals the stress response of the individual clownfish.

\section{Conclusion}\label{section5}
In this work, we proposed an estimation algorithm based on optical flow to calculate the velocity, magnitude, and turning angle for identifying the dispersion and motion behaviors of clownfish \textit{(Amphiprion bicinctus)} to enhance the welfare in ornamental fish aquaculture. Two tests based on the individual and group behaviors and morning versus afternoon behaviors were carried out on a database of two days of video streams of clownfish species in the Coastal and Marine Resources Lab at KAUST. The results showed that the estimated displacement revealed good performance in identifying clownfish's motion and dispersion characteristics. \textcolor{black}{We noted that the individual fish occurrence velocity's probability is higher than the group behavior, revealing a possible individual clownfish's stress activity level and a latent effect of the fish stocking density. However, these observations' results cannot be extrapolated to the clownfish population in general, as they do not include replicated pilots. Ultimately, the methods and algorithms developed in this paper for clownfish species serve as a technology test in a non-replicated pilot study of clownfish behavior. In addition, these test technology-oriented tools help provide practical baseline support for online predicting and monitoring particular behaviors in ornamental fish aquaculture.}
Finally, combining the proposed optical flow method and deep learning algorithm can help improve fish stress detection performance. For instance,  the optical flow algorithm can provide the neural networks' weights, enhancing key feature spaces. Hence, this will help evolve, for example, the feeding and water quality control policies by systematically improving their performances based on feedback. The proposed estimation scheme will be extended in future work to detect and identify feeding activity and stress responses simultaneously.
\textcolor{black}{Further, we will provide more investigations within an extended period of a sequential and spatial collection of records with separate fish tanks containing other groups of $16$ clownfish maintained in three $80$ L tanks with one, five, or ten individuals to support our findings.}

\bibliography{biblio}

\end{document}